 \documentclass[final,1p,times,twocolumn]{elsarticle}

\usepackage{amssymb}

\usepackage{graphicx}
\usepackage{amsfonts}
\usepackage[cmex10]{amsmath}
\usepackage{amssymb}
\usepackage{graphics} 
\usepackage{epsfig}
\usepackage{subfig}
\usepackage{color}
\usepackage{mathtools}
\usepackage{amsmath}
\usepackage{blindtext, graphicx}
\usepackage{listings}
\usepackage{booktabs}
\usepackage{amssymb}
\usepackage{pifont}
\usepackage{caption}
\usepackage[ruled]{algorithm2e}
\usepackage{algorithmic}
\newcommand{\cmark}{\ding{51}}%
\newcommand{\xmark}{\ding{55}}%

\journal{ELSEVIER Signal Processing}

\begin{document}

\begin{frontmatter}



\title{Evolutionary Simplicial Learning as a \\ Generative and Compact Sparse Framework for Classification}
\author{Yigit Oktar\fnref{label1}}\ead{Yigit.Oktar@gmail.com}
\fntext[label1]{Department of Computer Engineering, Izmir University of Economics, Izmir, Turkey.}
\author{Mehmet Turkan\fnref{label2}}\ead{Mehmet.Turkan@ieu.edu.tr}\ead[url]{http://people.ieu.edu.tr/en/mehmetturkan}
\fntext[label2]{Department of Electrical and Electronics Engineering, Izmir University of Economics, Izmir, Turkey. Corresponding author.}
\address{Department of Computer Engineering\\
Department of Electrical and Electronics Engineering\\
Izmir University of Economics\\
Izmir, Turkey}

\begin{abstract}
	Dictionary learning for sparse representations has been successful in many reconstruction tasks. Simplicial learning is an adaptation of dictionary learning, where subspaces become clipped and acquire arbitrary offsets, taking the form of simplices. Such adaptation is achieved through additional constraints on sparse codes. Furthermore, an evolutionary approach can be chosen to determine the number and the dimensionality of simplices composing the simplicial, in which most generative and compact simplicials are favored. This paper proposes an evolutionary simplicial learning method as a generative and compact sparse framework for classification. The proposed approach is first applied on a one-class classification task and it appears as the most reliable method within the considered benchmark. Most surprising results are observed when evolutionary simplicial learning is considered within a multi-class classification task. Since sparse representations are generative in nature, they bear a fundamental problem of not being capable of distinguishing two classes lying on the same subspace. This claim is validated through synthetic experiments and superiority of simplicial learning even as a generative-only approach is demonstrated. Simplicial learning loses its superiority over discriminative methods in high-dimensional cases but can further be modified with discriminative elements to achieve state-of-the-art performance in classification tasks.
\end{abstract}


\end{frontmatter}


\section{Introduction}
\label{sec:Intro}

Sparse representations have been proven to be very successful at restoration and reconstruction tasks such as compression, denoising, deblurring, inpainting and superresolution~\cite{5420029}. In essence, they aim at modeling the data/signal through concise linear combinations attained from an overcomplete basis or set of elements. This overcomplete set of elements is named as the \textit{dictionary} and it can either be carefully fixed (experimentally or analytically) or be adapted to the data at hand through learning~\cite{tosic2011dictionary}. Conventional nonconvex optimization of dictionary learning for sparse representations is given in Eqn.~(\ref{eq:sparse_rep}) as follows,
\begin{equation}
\mathop {\arg\min}\limits_{ {\bf A}, \{{\bf x}_i\}} \sum_i{\| {\bf y}_i-{\bf A}{\bf x}_i \|_2^2 }~~\text{subject to}~~\|{\bf x}_i\|_0 \leq q,~\forall i,
\label{eq:sparse_rep}
\end{equation}

\noindent where the matrix $\bf A$ is the designated overcomplete dictionary and ${\bf x}_i$ is the sparse representation vector of the data point ${\bf y}_{i}, \forall i$. While minimizing the reconstruction error of ${\bf y}_{i}$ over the dictionary $\bf A$, each sparse vector ${\bf x}_{i}$ can have a maximum $q$ number of nonzero components due to the strict $\ell_0$-norm constraint. In literature, there exist approximate iterative solutions (namely, \textit{sparse coding} and \textit{dictionary update}) to this highly nonconvex problem and its variants~\cite{7088631}.

In addition to reconstructive signal processing tasks, dictionary learning can also be employed in machine learning problems such as classification and clustering~\cite{7404063,OKTAR201820,oktar2019k}. At this point, it is proper to introduce one-class classification, as the fundamental form of the general classification problem, to bridge the gap between reconstructive signal processing and machine learning. Supervised machine learning in the form of classification inherently suggests the existence of more than one label. The concept of one-class learning, also known as one-class or unitary classification, emerges when there only exists a single label within the dataset, and one needs to discriminate it against all possible unseen labels~\cite{moya1996network}. It is actually a special case of binary classification where there is the ``in-class'' label and also the ``out-of-class'', but there is not any or enough number of ``out-of-class'' samples within the training dataset. Therefore, in the absence or weakness of the opposing class samples, conventional binary classification methods will have difficulties as they target the decision boundary in-between.

One-class learning methods can be categorized by the type of the targeted classifier model. There exist decision-boundary approaches which seek enclosing hyperspheres, hyperplanes or hypersurfaces in general~\cite{khan2014one}. These methods can adjust the level-of-detail through the usage of parametrized kernels to cope with the over- or under-fitting problem. On the other hand, graph-based methods try to fit a skeleton with-in data in a bottom-up manner. As an example, a minimum spanning tree model can be utilized as a one-class classifier~\cite{juszczak2009minimum}, in which the classification procedure relies on the distance to the tree. A generalization of graph-based approaches is attained through the concept of hypergraph, in which a hyperedge can now connect more than two data points or vertices. Hypergraph models not only allow custom but also lead the way to heterogeneous dimensionality. Such models are investigated in~\cite{wei2003hot,silva2008hypergraph}. As detailed in Sec.~\ref{sec:simplicial_intro}, simplicial learning through an extension of dictionary learning can be thought as the utmost generalization of the graph-based domain, in which vertices of a hypergraph can now move freely in space, taking the form of a simplicial.

By definition, an inner-skeleton method seeks a low and possibly heterogeneous dimensional piecewise linear model that expresses the data well in a compact manner. Most importantly, the dictionary learning concept can be categorized as an inner-skeleton method. However, the skeleton attained is not bounded in space but rather an infinite one, where each infinite linear bone is connected to all others at the origin. Technically speaking, a bone corresponds to a linear subspace of arbitrary dimensions. This conception will be indeed helpful when dictionary learning is considered within a multi-class classification framework. In its traditional multi-class formulations, the sparse representation based classifier models a separate dictionary for each distinct class through a data fidelity term together with an $\ell_p$-norm regularization constraint on sparse codes ($p=0$ or $1$ in general). Later, the test data is encoded sparsely and classified accordingly favoring the most reconstructive or representative dictionary~\cite{wei2013locality}. In the absence of other modifications, this form of sparse representation based classifier is known to be generative-only. The generative type approaches can create natural random instances of a class, in contrast to discriminative-only methods which focus on decision boundaries between classes.

In a simplistic manner, one can draw parallels between inner-skeleton and generative formulations which discard the existence of other classes; on the other hand, also between decision-boundary and discriminative approaches which need the existence of opposing classes. Not surprisingly, a method can be both generative and discriminative at the same time. Discrimination, in this sense, rises from the fact that while learning a dictionary (or a model) for a class, the data points from other classes are also taken into consideration, i.e., distance to those other points are to be maximized. Some examples of discriminative dictionary learning methods can be given as~\cite{mairal2009supervised,jiang2013label}.

\begin{figure*}[!t]%
	\centering
	\includegraphics[width=6cm]{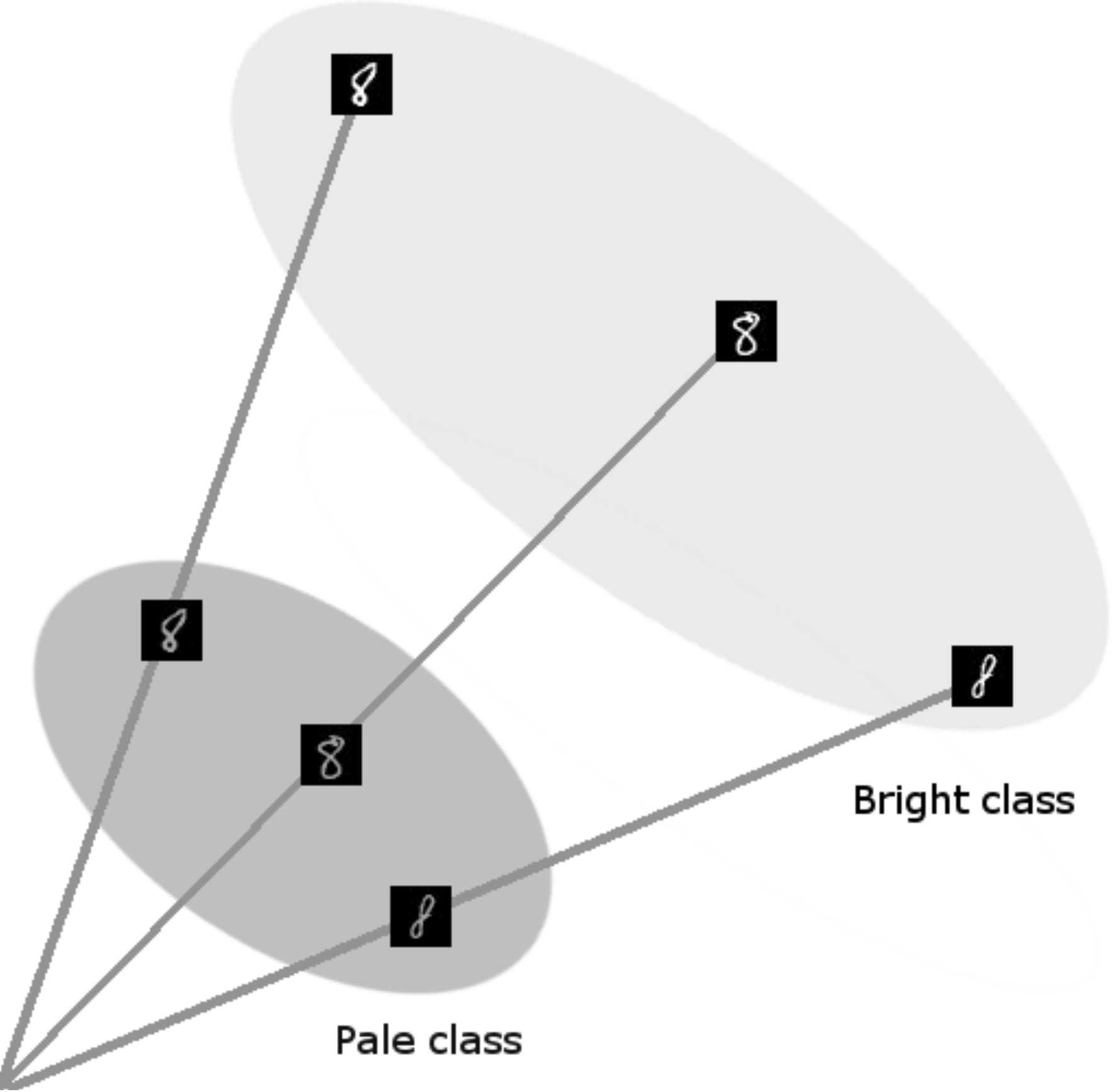}%
	\caption{Conventional dictionary learning is incapable of distinguishing intensity/magnitude, or more technically two classes within the same subspace.}%
	\label{fig:sdl_incapable}%
\end{figure*}

There is a subtle but crucial point that goes unnoticed in sparse representation based classifier applications and this forms the backbone of the proposed study in this paper. Corresponding to this upcoming point, XOr problem of neural networks dictates that a single layer perceptron is not capable of separating XOr inputs as only a single linear decision boundary is at hand. This has paved way to multilayer formulations that can solve linearly non-separable cases. A similar problem haunts dictionary learning methods silently. Consider the case as demonstrated in Fig.~\ref{fig:sdl_incapable}, in which there are two classes of digit $8$. ``Pale class'' includes pale images, while ``Bright class'' contains exactly the same images but they are brightened up. In technical terms, there are two opposing classes lying on the same subspace in the eyes of linear dictionary learning methods. No matter how much discriminative they are, traditional techniques will be incapable of totally distinguishing these two classes. In other words, dictionary learning in its conventional form is insensitive to intensity/magnitude and it will never be able to solve problems requiring intensity/magnitude distinction.

This study proposes a new dictionary learning framework for sparse representations through simplicials. While adapting conventional optimization constraints on sparse codes, the developed evolutionary simplicial learning algorithm leads to a strong generative approach. Experimental validation on different classification tasks demonstrates that this generative-only structure can successfully distinguish two different classes lying on the same subspace as an advantage, while there exist some shortcomings when its discriminative power is under consideration. Achieving state-of-the-art performance in most cases is highly possible through further modifications with
discriminative elements. The remaining part of this paper is organized as follows. Sec.~\ref{sec:simplicial_intro} introduces the basic concepts and mathematical foundations of simplicial learning as an extension to classical dictionary learning for sparse representations. Then, Sec.~\ref{sec:Evol} details the proposed simplicial learning algorithm by adopting an evolutionary approach with the appropriate fitness function to the problem. Sec.~\ref{sec:Exp} later reports experimental simulations over several datasets and illustrates the obtained results in different classification tasks. Finally, Sec.~\ref{sec:Conc} briefly concludes this study together with possible considerations which can be adapted to strengthen both theoretical and application aspects of the proposed framework.

\section{Simplicial Learning: An Extension of Dictionary Learning}
\label{sec:simplicial_intro}

\subsection{Definitions}
Dictionary learning optimization in Eqn.~(\ref{eq:sparse_rep}) basically tries to fit a union of subspaces to the data. Such subspaces are indeed infinite-extent and all crossing the origin without offsets, designated by the dictionary elements usually referred to as \textit{atoms}. Simplicial learning as an adaptation of dictionary learning aims instead at fitting bounded generic piecewise linear objects to the data. Table~\ref{table1} considers certain bounded generic piecewise linear objects. There are many not-equivalent formal definitions of the first construct, namely a \textit{polytope} to be discussed. This study strictly sticks with the definition that ``a polytope is an intact object which admits a simplicial decomposition.'' Hence, a polytope is made up of one or more simplices, whereas it is still in question that such simplices can be of different dimensions.

There are two possible ways to generalize the concept of polytope. In the first generalization, connectedness can be discarded leading to the fact that there is not a single object but multiple objects being considered at the same time. The second one allows the building-blocks namely simplices to have different dimensions, thus leading to heterogeneously dimensional objects. A formal name for such union of simplices is a \textit{simplicial complex}, but restricted self-intersections are imposed for a rigorous treatment. By definition, a simplicial complex is a set of simplices satisfying the following two conditions: (\textit{i}) every face of a simplex from this set is also in this set and (\textit{ii}) the non-empty intersection of any two simplices is a face of these two simplices. Losing a bit of formalism, utmost flexibility can be reached by allowing such objects to intersect each other and themselves in arbitrary ways, and such final construct is simply named as a \textit{simplicial} in the remaining part of this paper, to refer to an arbitrary union of simplices in the most general sense. For a more rigorous treatment of these definitions and related concepts, readers might refer to~\cite{munkres2018analysis}. 
\begin{table}[!t]
	\centering
	\renewcommand{\arraystretch}{0.5}
	\caption{Distinctions between the terms for generic objects.}
	\label{table1}
	\begin{tabular}{lcccc}
		\toprule
		& \begin{tabular}[c]{@{}l@{}}\footnotesize{May not}\\ \footnotesize{be intact}\end{tabular} & \begin{tabular}[c]{@{}l@{}} \footnotesize{Piecewise}\\ \footnotesize{linear} \end{tabular} & \begin{tabular}[c]{@{}l@{}} \footnotesize{Heterogeneous}\\ \footnotesize{dimensionality} \end{tabular} & \begin{tabular}[c]{@{}l@{}} \footnotesize{Arbitrary}\\ \footnotesize{intersections} \end{tabular}  \\  \midrule
		\footnotesize{Polytope} & \xmark & \cmark & \textbf{?} & \cmark \\
		\footnotesize{Simplicial complex} & \cmark & \cmark & \cmark & \xmark \\
		\footnotesize{Simplicial} & \cmark & \cmark & \cmark & \cmark \\ 
		\bottomrule
	\end{tabular}
\end{table}

\subsection{Related work}
Simplex and simplicial complex based data applications are becoming popular in literature as data analysis receives more and more topological considerations~\cite{luo2017learning,huang2015new,belton2018learning,tasaki2016simplex,patania2017topological}. Moreover, utilizing simplices for data applications is not a completely new idea from the perspective of sparse representations~\cite{wang2016recognizing,nguyen2013simplicial}.  Quite similarly, in this study an adaptation of sparse representations framework is chosen that casts a union of subspaces to a union of simplices. A rigorous mathematical formulation is detailed in the following.
\begin{figure*}%
	\centering
	\subfloat[Subspace]{{\includegraphics[width=4.5cm]{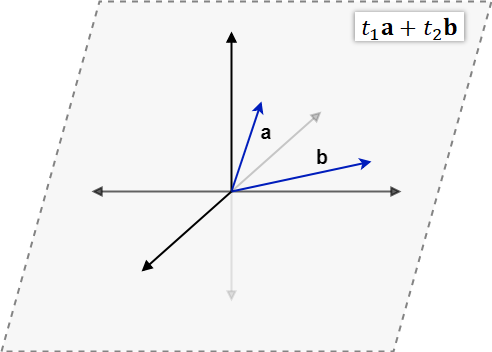} }}%
	\subfloat[Flat]{{\includegraphics[width=4.5cm]{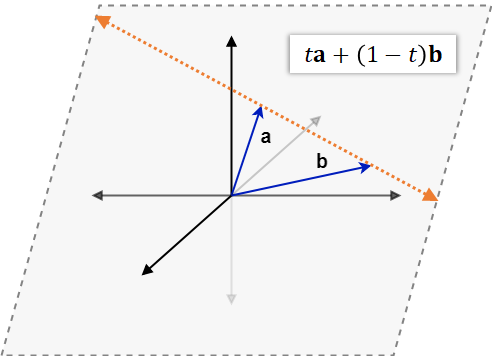} }}%
	\subfloat[Simplex]{{\includegraphics[width=4.5cm]{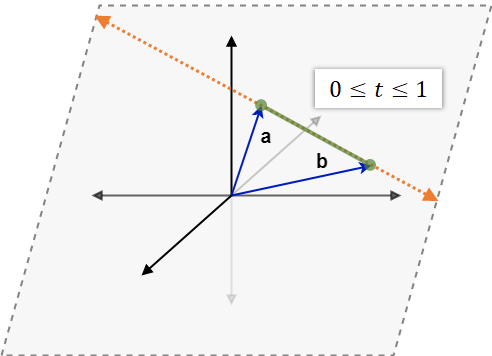} }}%
	\caption{A simple example of how additional constraints on sparse codes affect the solution of sparse representations. (a) The conventional sparsity constraint together with (b) sum-to-one ($t_1+t_2=1$) and (c) sum-to-one and non-negativity ($t_1+t_2=1$ and $t_1,t_2\geq0$) constraints.}%
	\label{sparse_Forms}%
\end{figure*}

\subsection{Mathematical formulation}
There are three necessary modifications to make a successful transition from the traditional dictionary learning formulation to simplicial learning. First of all, an additional sum-to-one constraint is needed on the sparse codes as noted in Eqn.~(\ref{flat_rep}) as follows,
\begin{equation}
\mathop {\arg\min}\limits_{ {\bf A}, \{{\bf x}_i\}} \sum_i{\| {\bf y}_i-{\bf A}{\bf x}_i \|_2^2 }~~\text{subject to}~~\|{\bf x}_i\|_0 \leq q~\wedge~{\bf 1}^{\text{T}}{\bf x}_i = 1,~\forall i,
\label{flat_rep}
\end{equation}

\noindent where $\bf 1$ denotes the column vector of ones, of appropriate size with the sparse vectors ${\bf x}_i$. Such modification casts $q$-dimensional subspaces into ($q-1$)-dimensional flats, a flat being a ($q-1$)-subspace with an arbitrary offset. A geometric explanation is illustrated in Fig.~\ref{sparse_Forms}(a-b) for the case when $q=2$. In this example, a subspace solution (i.e., an infinite-extent plane) of sparse representations is indeed reduced into a flat (i.e., an infinite-extent line) with an additional sum-to-one constraint on sparse codes.

In addition to above constraint, the second necessary modification is an additional non-negativity on sparse codes as noted in Eqn.~(\ref{simplex_rep}) as follows, 
\begin{equation}
\mathop {\arg\min}\limits_{ {\bf A}, \{{\bf x}_i\}} \sum_i{\| {\bf y}_i-{\bf A}{\bf x}_i \|_2^2 }~~\text{subject to}~~\|{\bf x}_i\|_0 \leq q~\wedge~{\bf 1}^{\text{T}}{\bf x}_i = 1~\wedge~{\bf 0} \leq {\bf x}_i,~\forall i,
\label{simplex_rep}
\end{equation}

\noindent where $\bf 0$ denotes the column vector of zeros, of appropriate size with the sparse vectors ${\bf x}_i$. Together with sum-to-one constraint, sparse codes are now restricted to $[0-1]$ range in magnitude and thus represented flat as an infinite-extent line turns into a simplex (i.e., a bounded line, line segment) as apparent in Figure~\ref{sparse_Forms}(b-c) for $q=2$. In the most generic sense, a simplex can be regarded as a bounded flat.

Note here that there is no any structural constraint on the sparse code patterns for the optimization problems in Eqns.~(\ref{eq:sparse_rep})-(\ref{simplex_rep}). In other words, all possible $q$-combinations of dictionary atoms are available for a $q$-sparse vector solution ${\bf x}_i$. Since most of these combinations are unnecessary for a given overcomplete dictionary, keeping a set of possible valid combinations (i.e., forcing certain patterns in sparse codes) will provide a more efficient and more compact representation. This finally leads to the concept of \textit{structured sparsity}, or \textit{group sparsity} in exact terms~\cite{yuan2006model,jacob2009group}, as a last modification on the road to simplicial learning.

While referring back to Sec.~\ref{sec:Intro}, when positional information is removed from a simplicial, the structure left then corresponds to a hypergraph, in which a hyperedge refers to a specific simplex within the simplicial. In relation to group sparsity, a hyperedge exactly corresponds to a group of atoms, hence a valid pattern of sparse codes. As a consequence, a set of groups/hyperedges, or more technically a hypergraph data structure needs to be kept to define the shape of the simplicial. This hypergraph structure will be denoted as ${\mathcal H} = \{h_{j}\}$ where $h_{j}$ designates the $j^{th}$ hyperedge referring to $j^{th}$ simplex within the simplicial. In accordance with this definition, simplicial learning with a structure imposed by ${\mathcal H}$ can be formulated in Eqn.~(\ref{simp_learn}) as follows, 
\begin{equation}
\begin{split}
&\mathop {\arg\min}\limits_{ {\bf A}, \{{\bf x}_i\}, \{h_j\}} \sum_i{\| {\bf y}_i-{\bf A}{\bf x}_i \|_2^2 }~~\text{subject to}\\
&\|{\bf x}_i\|_0 \leq q_j^*~\wedge~{\bf 1}^\text{T}{\bf x}_i = 1~\wedge~{\bf 0} \leq {\bf x}_i~\wedge~\left(k \notin h_j^* \rightarrow {\bf x}_i^{k}=0,~\forall k\right),~\forall i,
\end{split}
\label{simp_learn}
\end{equation}

\noindent where $h_j^*$ is the hyperedge indexing the closest simplex for the data point ${\bf y}_i$, $q_j^*=\lvert{h_j^*}\rvert$ denotes the dimension of that simplex, and the $\left(k \notin h_j^* \rightarrow {\bf x}_i^{k}=0,~\forall k\right)$ constraint ensures the group sparsity such that only the optimal group (i.e., hyperedge referring to the closest simplex) in ${\bf x}_i$ is to be filled and other entries which are represented as ${\bf x}_i^k$ shall all be zero. Note here that groups can be not only overlapping but also of different sizes, hence leading to heterogeneous dimensionality. In this final form, ${\mathcal H}$ needs to be learned together with ${\bf A}$ but a further careful consideration is needed over the compactness of the simplicial in return.

In summary, as is, the optimization in Eqn.~(\ref{simp_learn}) is highly ill-posed since there is no restriction on the number of simplices to be used or the dimensions of those simplices. One could even choose a very high-dimensional simplicial construct and zero-out the approximation error easily. Therefore, additional penalty terms need to be investigated based on the number and the dimensionality of simplices for a compact solution. Such a challenge appears to be highly combinatorial in nature and an evolutionary approach can be adopted after a careful consideration of an appropriate fitness function, as described and detailed in Sec.~\ref{sec:Evol}.

\section{Evolutionary Approach}
\label{sec:Evol}

To obtain an optimal or a suitable simplicial in a heuristic manner, certain number of simplicials are to compete against each other on instances of the same dataset. Basically, an evolutionary approach includes a suitable fitness function to guide this search process, and sub-procedures such as \textit{mutations} and \textit{breeding} to perform the actual search.

\subsection{The fitness function}

There are certain critical points to be carefully considered before designating the fitness function for the defined problem in this study. First of all, a straightforward optimization procedure for the number and the dimensionality of simplices will not be enough to attain a compact model desired. For example, consider that the data is distributed in the shape of a triangle with certain area. In this case, a triangle with the most compact area should be preferred as a targeted model. However, one could fit a triangle to this data with correct angles but excessive area. In such a case the dimensionality or the number of simplices indeed do not change. In conclusion, one needs also to take the volume, or more technically the content of the simplicials, besides considering the number and the dimensionality of simplices.

The content (or volume) of an arbitrary simplex can be calculated using Cayley-Menger determinant~\cite{li2015simplex}. Let $V$ be a $q$-dimensional simplex in $\mathbb{R}^N$, and $\bf B$ denote $(q+1)\times (q+1)$ distance matrix of vertices $\{v_1, v_2,...,v_{q+1}\}$ such that ${\bf B}_{ik} = \| v_{i}-v_{k} \|_{2}^{2}$. Then the content $C_{V}$ of $V$ is given in a relation in Eqn.~(\ref{cont}) as follows,
\begin{equation}
C_{V}^{2} = \frac{(-1)^{q+1}}{2^{q}(q!)^{2}}det(\hat{\bf B}),
\label{cont}
\end{equation}	

\noindent where $\hat{\bf B}$ is $(q+2)\times (q+2)$ matrix obtained from $\bf B$ by bordering it with a top row of $(0,1,...,1)$ and a left column of $(0,1,...,1)^\text{T}$.

Related with the content calculation here, another issue arises because of the allowed heterogeneous dimensionality in the optimization formula. The content of a line-segment (as an object) and a triangle (as an object) are incomparable in a general continuous setting since a triangle contains infinitely-many line-segments itself. To resolve this problem, an exponential term is introduced through \textit{an approximated cumulative discrete content} calculation of a simplicial as given in Eqn.~(\ref{app_content}) as follows,
\begin{equation}
\sum_{j=1}^{\lvert{\mathcal H}\rvert}{(1+C_{j})^{q_{j}}},
\label{app_content}
\end{equation}

\noindent where $\lvert{\mathcal H}\rvert$ denotes the number of hyperedges or equivalently the number of simplices, $C_{j}$ is the content of the $j^{th}$ simplex and $q_{j}$ is the dimension of that simplex. As a content $C_j<1$ would complicate the exponentiation used, $\left(1+C_j\right)$ is needed in the discrete approximation.

Having pinned down the above term which will be a component in the fitness function driving the evolutionary process, a fitness function candidate (in a minimization form) is given in Eqn.~(\ref{fit1}) as follows,
\begin{equation}
\sum_i{\| {\bf y}_i-{\bf A}{\bf x}_i \|_2^2 } + \alpha\sum_{j}{(1+C_{j})^{q_{j}}},
\label{fit1}
\end{equation}

\noindent where sum of squared error (SSE) used as the data fidelity term and approximated cumulative discrete content as to regulate the compactness of the representation. $\alpha$ denotes the regularization parameter controlling the contribution of the compactness prior on the solution.

While initially experimenting above fitness function, it is observed that the parameter $\alpha$ has a very broad optimality range, which changes drastically from dataset to dataset. This is due to the fact that there is a high dynamic range imbalance between two cumulative terms. Therefore, a variant of the defined fitness function is considered by transforming Eqn.~(\ref{fit1}) into the logarithmic scale in order to compress the dynamic range, leading to a more natural maximization setting formulated in Eqn.~(\ref{fitfin}) as follows,
\begin{equation}
\frac{\log_{10}\left(\frac{n}{\sum_i{\| {\bf y}_i-{\bf A}{\bf x}_i \|_2^2 } }\right)}{1+\beta\log_{10}\left({\gamma+\sum_{j}{(1+C_{j})^{q_{j}}}}\right)},
\label{fitfin}
\end{equation}

\noindent where $n$ denotes the number of data points and the parameter $\beta$ regulates over- or under-fitting. When $\beta=0$, the fitness function simply reduces to the data fidelity term favoring only for the reconstruction quality. Instead, a high $\beta$ value forces the simplicial to be compact. Empirical investigations suggest that a $\beta$ value around $0.05$ could be a global setting as it provides excellent results over all datasets considered in this study. The parameter $\gamma$ is fixed to $10$.

\subsection{Mutations and breeding}

First of all, it is important to note here that the hypergraph ${\mathcal H}$ is kept in the form of an incidence matrix of zeros and ones, where the row count corresponds to the number of simplices and the column count matches to the number of vertices or rather the number of atoms (columns) in the dictionary $\bf A$. Mutations can be easily applied on this binary matrix. In detail, there are four main processes that provide the background for evolution: (\textit{i}) increasing/decreasing the dimension of a simplex, (\textit{ii}) adding/removing a simplex, (\textit{iii}) subdividing a simplex and (\textit{iv}) adding/removing a vertex. All of these mutation operations are performed randomly without any optimality consideration.

As an additional tool to assist the searching process, breeding of two simplicials is also undertaken in which both dictionary elements and hypergraph structures of those two simplicials are split and then merged appropriately in order to create a new simplicial representative of two parents up to certain extent. Details of the breeding procedure are depicted in Alg.~\ref{alg_breed}. At first, hypergraph structures and the corresponding dictionary elements are extracted for these two simplicials $S_1 = \left(\mathcal{H}_1,{\bf A}_1\right)$ and $S_2 = \left(\mathcal{H}_2,{\bf A}_2\right)$. Then random submatrices $\mathcal{H}_{a}\in \mathcal{H}_{1}$ and $\mathcal{H}_{b}\in \mathcal{H}_{2}$ from each hypergraph are attained together with the corresponding columns of these dictionaries, contained in matrices ${\bf A}_a\in {\bf A}_1$ and ${\bf A}_b\in {\bf A}_2$. While vertices (atoms) are directly concatenated in ${\bf A}_{new}$ (line $7$), hypergraphs are concatenated in a disjoint manner in $\mathcal{H}_{new}$ (line $8$). In short, two subsimplicials are extracted and then grouped together in a disjoint manner to form a new simplicial $S_{new}$. Such tool can be suitably employed to exploit the underlying dimensionality of the dataset since these splitting and merging processes may lead child simplicials to acquire a properly representative data-dimensionality in a very fast manner, much faster than mutation processes to perform alone. Therefore, as a general observation, breeding determines the core dimensionality of the simplicial and mutations fine-tune the simplicial to the data.

\begin{algorithm}[!t]
	\caption{Breeding Algorithm}
	\begin{algorithmic}[1]
		\STATE $\left(\mathcal{H}_{1}, {{\bf A}_{1}}\right) \gets \text{get\_structure}(S_{1})$
		\STATE $\left(\mathcal{H}_{2}, {{\bf A}_{2}}\right) \gets \text{get\_structure}(S_{2})$
		\STATE $\mathcal{H}_{a}  \gets~\text{a random submatrix of}~\mathcal{H}_{1}$ 
		\STATE ${{\bf A}_{a}}  \gets~\text{the submatrix of}~{{\bf A}_{1}}~\text{corresponding to}~\mathcal{H}_{a}$ 
		\STATE $\mathcal{H}_{b}  \gets~\text{a random submatrix of}~\mathcal{H}_{2}$ 
		\STATE ${{\bf A}_{b}}  \gets~\text{the submatrix of}~{{\bf A}_{2}}~\text{corresponding to}~\mathcal{H}_{b}$ 	
		\STATE ${{\bf A}_{new}} \gets \begin{bmatrix} {{\bf A}_{a}} & {{\bf A}_{b}} \end{bmatrix}$
		\item[]
		\STATE ${\mathcal{H}_{new}} \gets  \begin{bmatrix} \mathcal{H}_{a} & 0 \\ 0 & \mathcal{H}_{b}  \end{bmatrix}$
		\item[]
		\STATE ${S_{new}} \gets \left({\mathcal{H}_{new}},{{\bf A}_{new}}\right)  $
	\end{algorithmic}
	\label{alg_breed}
\end{algorithm}

\subsection{Implementation details}

The algorithm to learn an evolutionary simplicial model on a set of data points $\left\{{\bf y}_i\right\}_{i=1}^n$ stored in the columns of a data matrix $\bf Y$ is given in Alg.~\ref{esl}. At first, the initial simplicial is to be generated from the given data points (line $1$). It is observed that choosing a single point (i.e., centroid of the dataset) as an initial simplicial is sufficient for low-dimensional problems. Through mutations and breeding processes, the initial simplicial takes an appropriate form in a fast manner since the search space is relatively small. However, a procedure involving the $k$-means algorithm~\cite{jain2010data} as a subroutine is employed to designate the initial simplicial for high-dimensional problems. In such cases, starting from a single point greatly slows down the process of evolution since the search space is quite large. Hence, an initialization based on $k$-means ensures that the starting simplicial is already a relatively fit one. A last point worth mentioning related to initialization here is that the initial simplicial $S$ should satisfy the condition that the numerator of Eqn.~(\ref{fitfin}) is positive, i.e., $\sum_i{\| {\bf y}_i-{\bf A}{\bf x}_i \|_2^2} < n$ to lead a meaningful evolution.

\begin{algorithm}[!t]
	\caption{Evolutionary Simplicial Learning (ESL) Algorithm}
	\begin{algorithmic}[1]
		\STATE $pop \gets \text{init\_pop}({\bf Y})$
		\WHILE{$not~converged$} 
		\STATE $pop \gets \text{mutations}(pop)$ 
		\STATE $pop \gets \text{breeding}(pop)$
		\FORALL{$S$ in $pop$} 
		\STATE ${\bf X} \gets \text{sparse\_coding}({\bf Y},S)$
		\STATE ${\bf A} \gets \text{dictionary\_update}({\bf Y},{\bf X})$
		\STATE $F \gets \text{fitness}(\bf{A},\mathcal{H})$
		\ENDFOR
		\STATE $pop \gets \text{sort and choose based on}~F~\text{values}$
		\ENDWHILE
		\STATE $S_{best} \gets pop(1)$
	\end{algorithmic}
	\label{esl}
\end{algorithm}

On line $6$, the algorithm performs the projection of data points $\left\{{\bf y}_i\right\}$ in $\bf Y$ onto each simplex of the simplicial $S$~\cite{duchi2008efficient,golubitsky2012algorithm} which basically corresponds to the sparse coding optimization. The closest simplex for the data point ${\bf y}_{i},~\forall i,$ is determined through the minimum approximation error acquired after projecting ${\bf y}_{i}$ onto each simplex. The positive barycentric coordinates of the projection points corresponding to the sparse codes are acquired, and then the necessary spots of the sparse representation matrix $\bf X$ is filled accordingly.

On line $7$, dictionary matrix $\bf A$ is updated using a direct least-squares solution. To optimize $\arg\min_{\bf A}\|{\bf Y}-{\bf A}{\bf X}\|_F^2$ by forcing its derivative to zero, the analytic solution is obtained with ${\bf A}={\bf Y}{\bf X}^+$ where ${\bf X}^+$ represents Moore-Penrose pseudo-inverse of {\bf X}. Note that there is no evolutionary process for learning $\bf A$, namely the vertices of the simplicial $S$. Instead, vertices are updated once exactly on this line at each iteration of the algorithm.

Finally, the surviving simplicials are determined based on the fitness scores they attain (line $10$). Experimental trials suggest that keeping the population size at $10$ is an efficient strategy, while an iteration count of $5$ is sufficient instead of a full convergence. Notice here that the parent simplicials are to be kept in the population pool when their fitness scores are higher than their children's.

\section{Experimental Results}
\label{sec:Exp}

The proposed method is tested in two phases of experiments to evaluate its classification capabilities. In the first experimental setup, the performance is evaluated in a one-class classification task for outlier detection. Datasets contain certain degree of outliers in such outlier detection problems, and methods learn models --agnostic of data labels-- in an unsupervised manner. In the second classification task, the performance of the proposed method is evaluated in a multi-class setting. At this stage, seven synthetic multi-class datasets are generated in addition to two handwritten digit recognition datasets. The synthetic datasets are special in that they contain cases which require intensity/magnitude distinction, especially very challenging for conventional dictionary learning methods.

\begin{table}[!t]
	\footnotesize
	\renewcommand{\arraystretch}{0.5}
	\caption{Information regarding the datasets used in outlier detection experiments.}
	\centering
	\begin{tabular}{@{}lcccc@{}}
		\toprule
		\textbf{Dataset} & \textbf{\#Samples} & \textbf{\#Dimensions} & \textbf{Outlier Ratio (\%)} \\ \midrule
		arrhythmia & $452$ & $274$ & $14.6018$ \\
		cardio & $1831$ & $21$ & $9.6122$   \\
		glass & $214$ & $9$ & $4.2056$  \\
		ionosphere & $351$ & $33$ & $35.8974$\\
		letter & $1600$ & $32$ & $6.2500$ \\
		lympho & $148$ & $18$ & $4.0541$  \\
		mnist & $7603$ & $100$ & $9.2069$   \\
		musk & $3062$ & $166$ & $3.1679$   \\
		optdigits & $5216$ & $64$ & $2.8758$\\
		pendigits & $6870$ & $16$ & $2.2707$ \\
		pima & $768$ & $8$ & $34.8958$\\
		satellite & $6435$ & $36$ & $31.6395$   \\
		satimage-2 & $5803$ & $36$ & $1.2235$\\
		shuttle & $49097$ & $9$ & $7.1511$ \\
		vertebral & $240$ & $6$ & $12.5000$  \\
		vowels & $1456$ & $12$ & $3.4341$ \\
		wbc & $378$ & $30$ & $5.5556$  \\
		\bottomrule
		\label{info_out}
	\end{tabular}
\end{table}

\subsection{Outlier detection}

In total $17$ benchmark datasets are taken from ODDS Library~\cite{Rayana:2016} for the one-class learning task. Information regarding these datasets in terms of number of samples, sample dimensionality and outlier percentages is summarized in Table~\ref{info_out} and interested readers might refer to~\cite{Rayana:2016} for details about each individual dataset. Using these benchmark datasets, a random $60\%$ to $40\%$ train-test set split is repeated for $10$ independent simulations and the mean Area Under The Curve (AUC) Receiver Operating Characteristics (ROC) results are reported in Table~\ref{out_res}.

The proposed Evolutionary Simplicial Learning (ESL) method is evaluated against an extensive outlier detection benchmark named as PyOD~\cite{zhao2019pyod}. The competing methods include Angle-based Outlier Detector (ABOD)~\cite{kriegel2008angle}, Clustering-based Local Outlier Factor (CBLOF)~\cite{he2003discovering}, Feature Bagging (FB)~\cite{lazarevic2005feature}, Histogram-based Outlier Score (HBOS)~\cite{goldstein2012histogram}, Isolation Forest (IForest)~\cite{liu2008isolation}, K Nearest Neighbors (KNN)~\cite{ramaswamy2000efficient}, Local Outlier Factor (LOF)~\cite{breunig2000lof}, Minimum Covariance Determinant (MCD)~\cite{hardin2004outlier}, One-class Support Vector Machine (OCSVM)~\cite{scholkopf2001estimating} and Principal Component Analysis (PCA)~\cite{PCAC} and one of the most recent results obtained in~\cite{weng2018multi} on the same benchmark (with an average of $20$ runs for each dataset).

\begin{table*}[!t]
	\tiny
	\renewcommand{\arraystretch}{1}
	\caption{Mean AUC ROC results from $10$ independent simulations for outlier detection on various datasets.}
	\centering
	\begin{tabular}{@{}lccccccccccccc@{}}
		\toprule
		\textbf{Dataset} & \textbf{ABOD} & \textbf{CBLOF} & \textbf{FB} & \textbf{HBOS} & \textbf{IForest} & \textbf{KNN} & \textbf{LOF} & \textbf{MCD} & \textbf{OCSVM} & \textbf{PCA} & \textbf{~\cite{weng2018multi}} & \textbf{ESL}  \\ \midrule
		arrhythmia & $0.769$ & $0.784$ & $0.778$ & $0.822$ & $0.801$ & $0.786$ & $0.779$ & $0.779$ & $0.781$ & $0.782$ & $0.801$ & $\bf 0.826$  \\
		cardio & $0.569$ & $0.928$ & $0.587$ & $0.835$ & $0.921$ & $0.724$ & $0.574$ & $0.814$ & $0.935$ & $0.950$ & $\bf 0.969$ & $0.884$  \\
		glass & $0.795$ & $0.850$ & $0.873$ & $0.739$ & $0.757$ & $0.851$ & $0.864$ & $0.790$ & $0.632$ & $0.675$ & - & $\bf 0.876$  \\
		ionosphere & $0.925$ & $0.813$ & $0.873$ & $0.561$ & $0.850$ & $0.927$ & $0.875$ & $\bf 0.956$ & $0.842$ & $0.796$ & $0.911$ & $0.851$  \\
		letter & $\bf 0.878$ & $0.507$ & $0.866$ & $0.593$ & $0.642$ & $0.877$ & $0.859$ & $0.807$ & $0.612$ & $0.528$ & - & $0.776$  \\
		lympho & $0.911$ & $0.973$ & $0.975$ & $\bf 0.996$ & $0.994$ & $0.975$ & $0.977$ & $0.900$ & $0.976$ & $0.985$ & $0.987$ & $0.984$  \\
		mnist & $0.782$ & $0.801$ & $0.721$ & $0.574$ & $0.816$ & $0.848$ & $0.716$ & $0.867$ & $0.853$ & $0.853$ & $\bf 0.929$ & $0.803$  \\
		musk & $0.184$ & $0.988$ & $0.526$ & $\bf 1.000$ & $\bf 1.000$ & $0.799$ & $0.529$ & $\bf 1.000$ & $\bf 1.000$ & $\bf 1.000$ & $\bf 1.000$ & $0.972$  \\
		optdigits & $0.467$ & $0.509$ & $0.443$ & $\bf 0.873$ & $0.725$ & $0.371$ & $0.450$ & $0.398$ & $0.500$ & $0.509$ & - & $0.746$  \\
		pendigits & $0.688$ & $0.949$ & $0.460$ & $0.924$ & $0.944$ & $0.749$ & $0.470$ & $0.834$ & $0.930$ & $0.935$ & $0.938$ & $\bf 0.951$  \\
		pima & $0.679$ & $\bf 0.735$ & $0.624$ & $0.700$ & $0.681$ & $0.708$ & $0.627$ & $0.675$ & $0.622$ & $0.648$ & - & $0.626$  \\
		satellite & $0.571$ & $0.669$ & $0.557$ & $0.758$ & $0.702$ & $0.684$ & $0.557$ & $\bf 0.803$ & $0.662$ & $0.599$ & $0.750$ & $0.705$  \\
		satimage-2 & $0.819$ & $0.992$ & $0.457$ & $0.980$ & $0.995$ & $0.954$ & $0.458$ & $0.996$ & $\bf 0.998$ & $0.982$ & $0.976$ & $0.995$  \\
		shuttle & $0.623$ & $0.627$ & $0.472$ & $0.986$ & $\bf 0.997$ & $0.654$ & $0.526$ & $0.990$ & $0.992$ & $0.990$ & $0.994$ & $0.992$  \\
		vertebral & $0.426$ & $0.349$ & $0.417$ & $0.326$ & $0.391$ & $0.382$ & $0.408$ & $0.391$ & $0.443$ & $0.403$ & $\bf 0.580$ & $0.413$  \\
		vowels & $0.961$ & $0.586$ & $0.943$ & $0.673$ & $0.759$ & $\bf 0.968$ & $0.941$ & $0.808$ & $0.780$ & $0.603$ & - & $0.881$  \\
		wbc & $0.905$ & $0.923$ & $0.933$ & $\bf 0.952$ & $0.931$ & $0.937$ & $0.935$ & $0.921$ & $0.932$ & $0.916$ & - & $0.924$  \\
		\hline
		MEAN & $0.703$ & $0.764$ & $0.677$ & $0.782$ & $0.818$ & $0.776$ & $0.679$ & $0.808$ & $0.793$ & $0.774$ & n/a & $\bf 0.835$  \\
		STDEV & $0.210$ & $0.195$ & $0.203$ & $0.192$ & $0.164$ & $0.182$ & $0.199$ & $0.179$ & $0.183$ & $0.197$ & n/a & $\bf 0.152$ \\ \bottomrule
		\label{out_res}
	\end{tabular}
\end{table*}

Last two rows of Table~\ref{out_res} illustrate the mean AUC ROC results over all datasets and their standard deviations. ESL not only presents the best average AUC ROC performance among all methods in the benchmark but also has the least standard deviation. One can conclude that it is the most reliable method among considered techniques for this performance measure. Moreover, ESL shows top AUC ROC performance in three datasets. However, additional tests show that it does not have a noticeable advantage in Precision at n (P@n) performance.

\subsection{Multi-class classification}

\begin{figure*}[!t]%
	\centering
	\subfloat[Cluster-in-Cluster]{{\includegraphics[width=4.5cm]{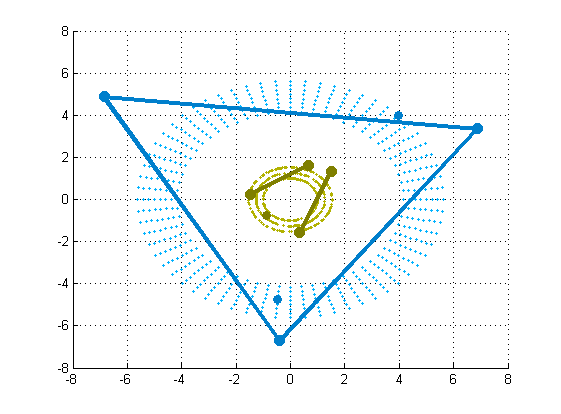} }}%
	\subfloat[Two-Spirals]{{\includegraphics[width=4.5cm]{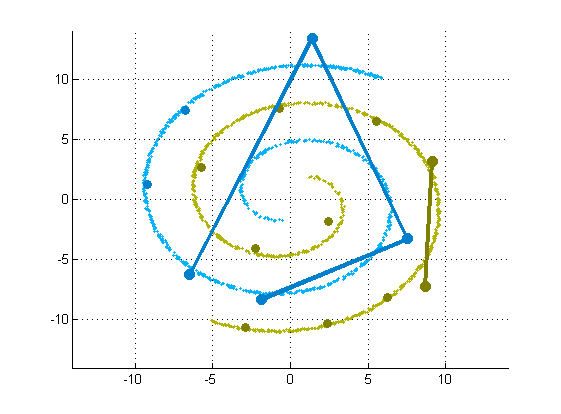} }}%
	\subfloat[Half-Kernel]{{\includegraphics[width=4.5cm]{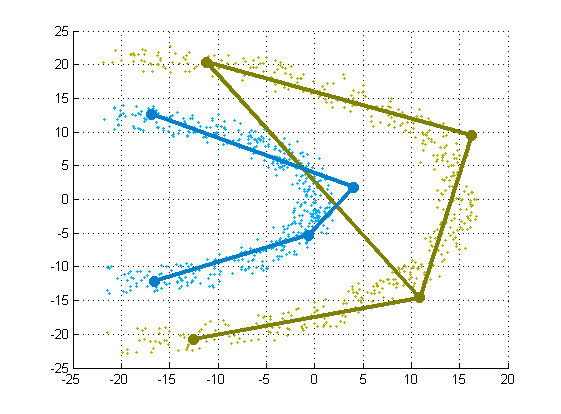} }}%
	
	\subfloat[Crescent\&Full-moon]{{\includegraphics[width=4.5cm]{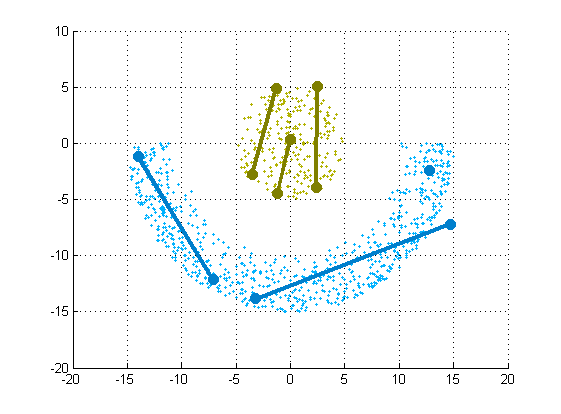} }}%
	\subfloat[Corners]{{\includegraphics[width=4.5cm]{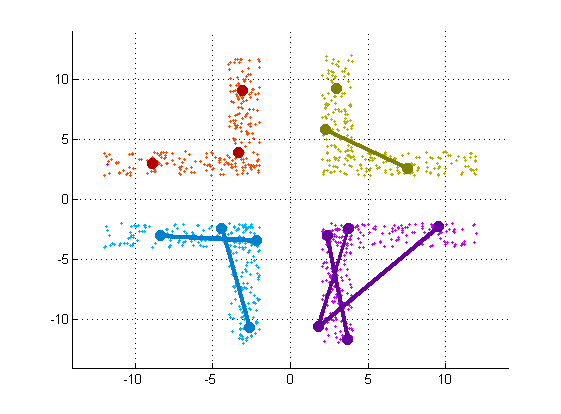} }}%
	\subfloat[Outliers]{{\includegraphics[width=4.5cm]{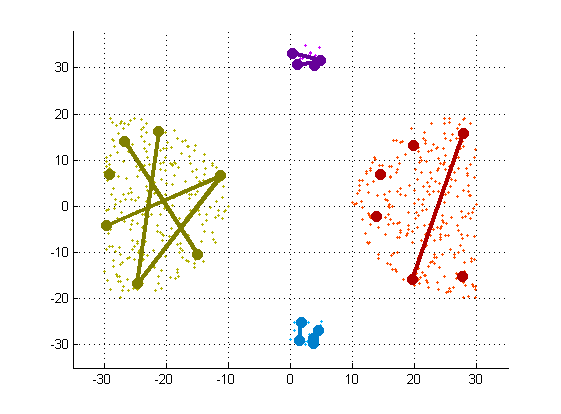} }}%
	\caption{Examples of learned simplicial models on six synthetic datasets. Best visualized in color.}%
	\label{fig:syn_Res}
\end{figure*}

For the multi-class classification task, six challenging synthetic datasets are generated by following the procedures in~\cite{noauthor_6_nodate} and these datasets are depicted in Fig.~\ref{fig:syn_Res}. Four of these datasets (namely, \textit{Cluster-in-Cluster}; \textit{Two-Spirals}; \textit{Half-Kernel} and \textit{Crescent\&Full-moon}) contain binary classification tasks while the remaining two of them (\textit{Corners} and \textit{Outliers}) consist of four-class classification problems. In addition, a synthetically altered dataset (named as MNIST$8$) is included in the experimental setup, in which all samples of the digit $8$ from the original MNIST~\cite{MNIST} are designated as the ``Bright class'' while a new ``Pale class'' is generated from all these original samples by dimming with a scale of $0.25$ according to the previous discussion related to Fig.~\ref{fig:sdl_incapable}.

\begin{table*}[!t]
	\caption{Classification success rates of different dictionary learning methods for six synthetic datasets, and for the proposed binary MNIST$8$ problem (last row).}
	\centering
	\tiny
	\renewcommand{\arraystretch}{1}
	\begin{tabular}{lcccccccc}
		\toprule
		\textbf{Dataset} & \textbf{SRC} & \textbf{LCKSVD1} & \textbf{LCKSVD2} & \textbf{DLSI} & \textbf{FDDL} & \textbf{DLCOPAR} & \textbf{LRSDL} & \textbf{ESL} \\ \midrule
		Cluster-in-Cluster & $52.77$ & $50.99$ & $54.55$ & $43.68$ & $55.53$ & $67.98$ & $45.45$ & $\bf 88.14$ \\
		Two-Spirals & $49.30$ & $41.50$ & $71.30$ & $52.30$ & $53.70$ & $51.10$ & $59.70$ & $\bf 80.22$ \\
		Half-Kernel & $63.80$ & $64.40$ & $65.60$ & $51.60$ & $58.00$ & $62.80$ & $64.80$ &  $\bf 93.65$ \\ 
		Crescent\&Full-moon & $75.00$ & $82.60$ & $78.00$ & $55.60$ & $64.40$ & $64.20$ & $85.60$  & $\bf 99.80$ \\ 
		Corners & $91.00$ & $25.00$ & $44.80$ & $27.80$ & $29.60$ & $29.20$ & $27.80$ & $\bf 97.50$ \\ 
		Outliers & $51.33$ & $43.33$ & $80.00$ & $52.33$ & $75.67$ & $53.33$ & $99.27$ & $\bf 100.00$ \\
		MNIST8 & $50.00$ & $50.00$ & $50.00$ & $50.00$ & $75.45$ & $50.05$ & $63.24$ & $\bf 99.05$ \\ \bottomrule
		\label{class_res_syn}
	\end{tabular}
\end{table*}

The proposed ESL algorithm in this setup is compared against Sparse Representation-based Classification (SRC)~\cite{wright2008robust}, Label Consistent K-SVD (LCKSVD1 and LCKSVD2)~\cite{jiang2013label}, Dictionary Learning with Structured Incoherence (DLSI)~\cite{ramirez2010classification}, Fisher Discrimination Dictionary Learning (FDDL)~\cite{yang2011fisher}, Dictionary Learning for Commonality and Particularity (DLCOPAR)~\cite{kong2012dictionary} and Low-rank Shared Dictionary Learning (LRSDL)~\cite{vu2016learning,vu2017fast}. Experimental results in terms of classification success rates are presented in Table~\ref{class_res_syn}. It is apparent that ESL easily outperforms all considered dictionary learning methods over all cases. This should not be a surprising result since all utilized synthetic datasets require intensity/magnitude distinction to various extents. On the other hand, some discriminative methods such as LCKSVD2, FDDL and LRSDL undergo meaningful learning (i.e., better than random) over some datasets. This observation leads to an important conclusion that discriminative modifications may alleviate insensitivity to intensity to a certain degree.

Fig.~\ref{fig:syn_Res} depicts examples of learned simplicial models on six synthetic datasets. As it can be observed clearly, simplicials are bounded and they are composed of simplices (i.e., points and line-segments in these cases) with arbitrary offsets, providing an advantage over unbounded and without-offset dictionary learning models in all these classification tasks.

\begin{table*}[!t]
	\caption{Classification error rates of various methods on handwritten digit datasets, USPS and MNIST. ESL appears as a superior generative method, nearly performing at the capacity of discriminative Gaussian SVM on both datasets.}
	\centering
	\tiny
	\renewcommand{\arraystretch}{1}
	\begin{tabular}{@{}lccccc|cccccc@{}}
		\multicolumn{11}{ c }{\textbf{Generative-only}~~~~~~~~~~~~~~~~~~~~~~~~~~~~~~~~~~~~~~~~~~~~~\textbf{Discriminative}}\\
		\toprule
		\textbf{Dataset} & \textbf{SDL-G} & \textbf{TDDL-G} & \textbf{LLC} & \textbf{LDL} & \textbf{ESL} & \textbf{KNN} & \textbf{SVM-Gauss} & \textbf{SDL-D} & \textbf{FDDL} & \textbf{TDDL-D} \\ \midrule
		USPS & $6.67$ & $4.58$ & $4.48$ & $\bf{3.79}$ & $4.31$ & $5.2$ & $4.2$ & $3.54$ & $3.69$ & $\bf{2.84}$ \\
		MNIST & $3.56$ & $2.36$ & - & - & $\bf{1.85}$ & $5.0$ & $1.4$ & $1.05$ & - & $\bf{0.54}$ \\ \bottomrule
	\end{tabular}
	\label{last_class}
\end{table*}

\textit{Digit Classification}: In most of the practical pattern recognition applications, the pattern or rather the direction of the feature vector utilized plays an important role on the success rate. For instance, a ``star pattern'' is a ``star pattern'' no matter how much bright or pale it is. Therefore, the advantage of simplicial learning over dictionary learning is expected to diminish in some real-world applications. This is observable in digit classification experiments featuring USPS~\cite{USPS} and MNIST datasets as reported in Table~\ref{last_class}. In this set of experiments, ESL is compared to classification methods including Supervised Dictionary Learning~\cite{mairal2009supervised} with generative training (SDL-G) and with discriminative learning (SDL-D), Task-driven Dictionary Learning~\cite{mairal2011task}: unsupervised (TDDL-G) and supervised (TDDL-D), FDDL, KNN, Gaussian SVM, Locality-constrained Linear Coding (LLC)~\cite{wang2010locality} and Locality-sensitive Dictionary Learning (LDL)~\cite{wei2013locality}. LLC and LDL methods have the sum-to-one constraint on sparse codes, therefore they learn spaces with arbitrary offsets but learned models are still not bounded (without the non-negativity constraint).

As apparent from Table~\ref{last_class}, ESL appears to be a successful generative-only method which performs nearly at the capacity of Gaussian SVM (i.e., a well-known and widely used discriminative classifier). However, it cannot outperform discriminative dictionary learning methods such as FDDL and TDDL-D in these datasets. A final note is that ESL can also be modified through discriminative elements. Discriminative methods SDL-D and TDDL-D have a $1.5-2\%$ advantage over their generative counterparts SDL-G and TDDL-G. Hence, a successful discriminative version of ESL can then be projected to reach state-of-the-art, an estimation open to discussion or further investigation.

\section{Discussion and Conclusion}
\label{sec:Conc}

Dictionary learning through simplicials is more flexible than classical dictionary learning models since simplices are bounded and freely positioned in space. The proposed sparsity based evolutionary structure, called ESL is highly applicable if the characteristics of the problem at hand requires such successful localized models. In this study, a global fitness function is employed and there is no restriction on the local fitness of each individual simplex within the simplicial. If the local fitness of each simplex is considered and optimized individually, the resulting simplicial model might be in a more compact form. For example, the unnecessary simplex of green simplicial in Fig.~\ref{fig:syn_Res}(c) would most probably be eliminated as it does not have any local fitness, thus lead to an increased accuracy of classification. Another point worth mentioning here is that the employed fitness function in Eqn.~(\ref{fitfin}) is reminiscent of Poisson distribution, in a multidimensional form~\cite{belyaev1988multidimensional}. Hence, other probabilistic considerations and also discriminative elements can be adapted to strengthen both theoretical and application aspects of the proposed framework.

As exemplified in this paper, simplicial learning can successfully address some weak points of conventional dictionary learning for the considered machine learning problems; it is a promising approach inherently capable of performing signal processing tasks and can become a general machine learning tool with many application domains.





\bibliographystyle{elsarticle-num}
\bibliography{refs}







\end{document}